\documentclass[11pt,a4paper]{article}

\usepackage[acceptedWithA]{tacl2021v1}
\usepackage{times}
\usepackage{latexsym}
\usepackage[T1]{fontenc}
\usepackage[utf8]{inputenc}
\usepackage{microtype}
\usepackage{inconsolata}
\usepackage{booktabs}
\usepackage{amsmath}
\usepackage{amssymb}
\usepackage{graphicx}
\usepackage{url}
\usepackage{float}

\newcommand{\fitcol}[1]{%
  \resizebox{\ifdim\width>\columnwidth \columnwidth\else \width\fi}{!}{#1}}
\newcommand{\fitwidth}[1]{%
  \resizebox{\ifdim\width>\textwidth \textwidth\else \width\fi}{!}{#1}}

\title{Learning to Fuse LLMs with Ontology Rankers for Rare-Disease Diagnosis}

\author{%
  \bfseries
  Zhaoyang Jiang\textsuperscript{1} \quad
  Xuanqi Peng\textsuperscript{1} \quad
  Fei Teng\textsuperscript{1} \quad
  Yunsoo Kim\textsuperscript{1} \quad\\[0.3em]
  \bfseries
  Zhizhong Fu\textsuperscript{3} \quad
  Jiacong Mi\textsuperscript{2} \quad
  Zicheng Li\textsuperscript{4} \quad
  Honghan Wu\textsuperscript{1}\thanks{Corresponding author.} \\[0.5em]
  \normalfont
  \textsuperscript{1}School of Health \& Wellbeing, University of Glasgow, Glasgow, UK \\
  \textsuperscript{2}Department of Respiratory and Critical Care Medicine, Shanghai Sixth People's Hospital, \\
  Shanghai Jiao Tong University School of Medicine, Shanghai, China \\
  \textsuperscript{3}School of Life Science and Technology, \\
  University of Electronic Science and Technology of China, Chengdu, China \\
  3167645J@student.gla.ac.uk,\quad
  Honghan.Wu@glasgow.ac.uk,
}

\begin{document}
\maketitle

\begin{abstract}
Ontology rankers remain useful for rare-disease diagnosis because each candidate can be traced to
matched patient phenotypes. Large language models (LLMs) can generate differential diagnoses from
the same patient description, but their predictions lack an equally clear evidence trail. Rather
than asking which system should replace the other, we ask whether an LLM can improve the ranker
without giving up its evidence. Our behavior-based fusion model examines the two ranked lists, their
agreement, and the ontology support behind each candidate, and learns how much to rely on each
system for the individual case. Before comparison, we remove a documented test-set leakage pathway
caused by benchmark cases and ontology annotations being derived from the same publications. Across
eight open LLMs, fusion improves Phenomizer Recall@1 by 7.86 percentage points on
Phenopacket Store and 20.18 points on RAMEDIS. When paired with
DeepSeek-V4-Flash through an API, a fusion model trained only on the other LLMs improves Recall@1
from 0.1657 to 0.2176, a 5.19-point gain,
without retraining. For 90.8\% of correct fused diagnoses, the disease retains
candidate-level ontology evidence that can be inspected. These results show that LLMs can
strengthen an established diagnostic tool without discarding the structured evidence that makes it
useful.
\end{abstract}

\section{Introduction}
\label{sec:intro}

Rare-disease diagnosis often begins with a list of findings rather than a familiar disease name.
Phenotype tools such as Exomiser~\cite{exomiser2014}, LIRICAL~\cite{lirical2020}, and
Phenomizer~\cite{phenomizer2009} compare those findings with disease profiles in the Human
Phenotype Ontology (HPO)~\cite{hpo2021} and return a ranked differential diagnosis. Their value is
not limited to the ranking. A clinician can inspect the HPO terms supporting each candidate and,
for Phenomizer, the statistical strength of the match. This evidence remains important in clinical
decision support, where clinicians need to examine the basis for a recommendation
~\cite{amann2020explainability}.

The arrival of LLMs has made it possible to produce a differential diagnosis directly from clinical
text or HPO terms, drawing on knowledge beyond a fixed curated resource. Yet the strongest
structured comparisons have favored classical tools. A large systematic benchmark reports
Exomiser ahead of every one of seven LLMs across accuracy metrics and clinical
subgroups~\cite{ejhg2026benchmark}, and prompt-only LLMs also trail established phenotype pipelines
in related rare-disease gene prioritization~\cite{lamarrvel2025}. LLM output introduces a second
concern. Medical models can elaborate fabricated clinical details~\cite{omar2025adversarial}, the
medical hallucination literature identifies factual reliability as a persistent
problem~\cite{zhu-etal-2025-trust}, and verbal confidence systematically overstates diagnostic
reliability~\cite{savage2025jamia}. A bare LLM ranking also lacks the independently computed HPO
matches and statistical score supplied by an ontology tool. On the surface, classical systems seem
to offer both greater accuracy and a more auditable decision path.

These findings establish the continuing strength of classical pipelines, but they do not determine
how much a ranker backed by the HPO annotation database (HPOA) benefits from the construction of a
particular benchmark. Phenopackets are structured patient records, many curated from published case
reports. HPOA is curated from the same literature. The same paper can therefore supply
both a test patient's findings and the phenotype evidence attached to that patient's confirmed
disease in the knowledge base. We call this \emph{publication-source overlap}.

Because both resources record publication provenance, we remove any phenotype relationship whose
only supporting source is the test case's paper before rerunning the unchanged ontology ranker.
After this correction, the ranker and LLM show complementary strengths rather than a simple winner.

This motivates a different role for the LLM. It need not replace the ontology tool, because the two
systems fail for different reasons and can supply what the other lacks. Recent work already shows
the value of this combination. LA-MARRVEL improves a phenotype-driven gene ranker with
language-aware reranking~\cite{lamarrvel2025}, while DeepRare integrates LLM reasoning with
specialized tools and traceable external evidence~\cite{deeprare2026}. Most directly,
\citet{elmofty-leser-2026-retrieval} find complementary correct diagnoses from ontology retrieval
and unrestricted LLM generation on every rare-disease benchmark they test. Their experiments also
expose the unresolved step. They call cases whose correct diagnosis falls outside the retriever's
candidate set non-retrievable, a hard ceiling for any method using that pool. They explicitly leave
a learned router using case characteristics as the next step. Complementary answers are available,
but identifying which one to use on a new case remains unresolved.

We learn a case-level decision rule. The LLM diagnoses independently rather than reordering the
ontology list, and a small behavior-based gate assigns case-specific weights before combining the
union of both outputs. It observes only their
behavior: how clearly the leading diagnoses stand out, whether the lists agree, how well patient
findings match the candidates, and how much ontology knowledge is available for them. The gate
receives neither model identity nor a model-specific representation. During training, we exclude
the target LLM together with every other model from its backbone family. It can therefore be
applied to a newly introduced LLM without first collecting labels for that model, allowing the
pipeline to adopt a stronger diagnostic model while retaining access to candidate-level ontology
evidence.

Code is available at \url{https://github.com/Anonymous-Awesome-Submissions/PhenoGate}.

\section{Related Work}

Ontology-based rare-disease diagnosis represents patient findings with the Human Phenotype
Ontology (HPO)~\cite{hpo2021}. Phenomizer introduced semantic-similarity disease ranking with an
empirical significance estimate~\cite{phenomizer2009}; Exomiser combined phenotype matching with
variant prioritization~\cite{exomiser2014}; and LIRICAL expressed phenotype evidence through
likelihood ratios~\cite{lirical2020}. Together, these systems established a diagnostic pipeline
grounded in curated disease profiles and inspectable HPO matches. Their continued strength in a
recent comparison with LLMs shows that they remain active diagnostic systems rather than merely
historical baselines~\cite{ejhg2026benchmark}.

LLM research first tested whether pretrained knowledge could recover diseases or genes directly
from phenotype descriptions, with RareBench and RareArena extending evaluation across diseases and
model families~\cite{rarebench2024,rarearena2025}. Later systems increasingly combined language
models with structured resources. LA-MARRVEL refines candidates from a phenotype-based gene
ranker~\cite{lamarrvel2025}, whereas DeepRare coordinates LLM reasoning with specialist tools and
external evidence~\cite{deeprare2026}. \citet{elmofty-leser-2026-retrieval} directly compare
ontology retrieval with unrestricted LLM diagnosis and find that the two solve complementary
cases. The literature has consequently moved from replacement toward hybrid diagnosis through
candidate reranking, tool-using agents, and parallel prediction.

Hybrid diagnosis also draws on rank aggregation and model selection. Information retrieval
developed score-, rank-, and probability-based fusion, including CombMNZ, Borda-fuse, Bayes-fuse,
ProbFuse, reciprocal-rank fusion, and Rank-Biased Centroids~\cite{foxshaw1994,aslam2001metasearch,
lillis2006probfuse,rrf2009,rbc2017}; later work studied learned score combinations for hybrid
retrieval~\cite{bruch2024fusion}. A parallel line learns when to accept one system or defer to
another, from confidence-based cascades and clinical deferral to preference-supervised and
label-free LLM routing~\cite{jitkrittum2023cascade,l2dclinical2026,routellm2025,smoothie2024}.
These studies replace a single rule for all inputs with input-dependent selection.

Benchmark research has meanwhile expanded its view of contamination. Work on LLM evaluation
primarily examines whether test examples appeared in pretraining data~\cite{balloccu2024leak};
prospective clinical benchmarks reduce that risk with newly collected cases~\cite{liveclin2026}.
For systems that retrieve external information, leakage can instead occur at inference time when
search exposes benchmark answers~\cite{searchcontam2026}. Open-domain question answering likewise
treats the retrievable evidence collection as part of the evaluated task
~\cite{kwiatkowski2019natural,lee2019latent}. Collectively, this literature shows that benchmark
validity depends not only on model training data, but also on what information the evaluation
pipeline makes available during inference.

\section{Correcting Publication-Source Overlap}
\label{sec:contam}

Consider a paper that reports one or more patients with disease $D$. It may be curated both as a
test case whose answer is $D$ and as knowledge-base entries associating $D$ with findings described
in the paper. A phenotype ranker then compares a test patient's findings with a disease profile
supported by the same document. HPOA records the publication supporting each relationship, making
this reuse of the source document directly observable. We say that a case has
\emph{publication-source overlap} when an annotation supporting its gold disease cites the case's
source publication, and call an annotation \emph{source-exclusive} when that publication is its
only recorded support.

For a case with source publication $p$, our leave-one-publication-out (LOPO) evaluation removes
each entry $e$ that links a disease to a phenotype when $\mathrm{refs}(e)=\{p\}$. Entries corroborated by any
other publication remain, as do entries without recorded provenance. We then rebuild each disease
profile by adding the HPO ancestors of its remaining annotations and rerun the unchanged ranker over
all candidate diseases. Information content is held fixed between the original and LOPO conditions,
so the comparison changes only the case-specific evidence derived exclusively from $p$.
Phenopacket Store records a source publication for each case, and HPOA records sources for
individual annotations, enabling this publication-level join.

\section{Ontology and LLM Fusion}
\label{sec:method}

\begin{figure*}[t]
\centering
\includegraphics[width=\textwidth]{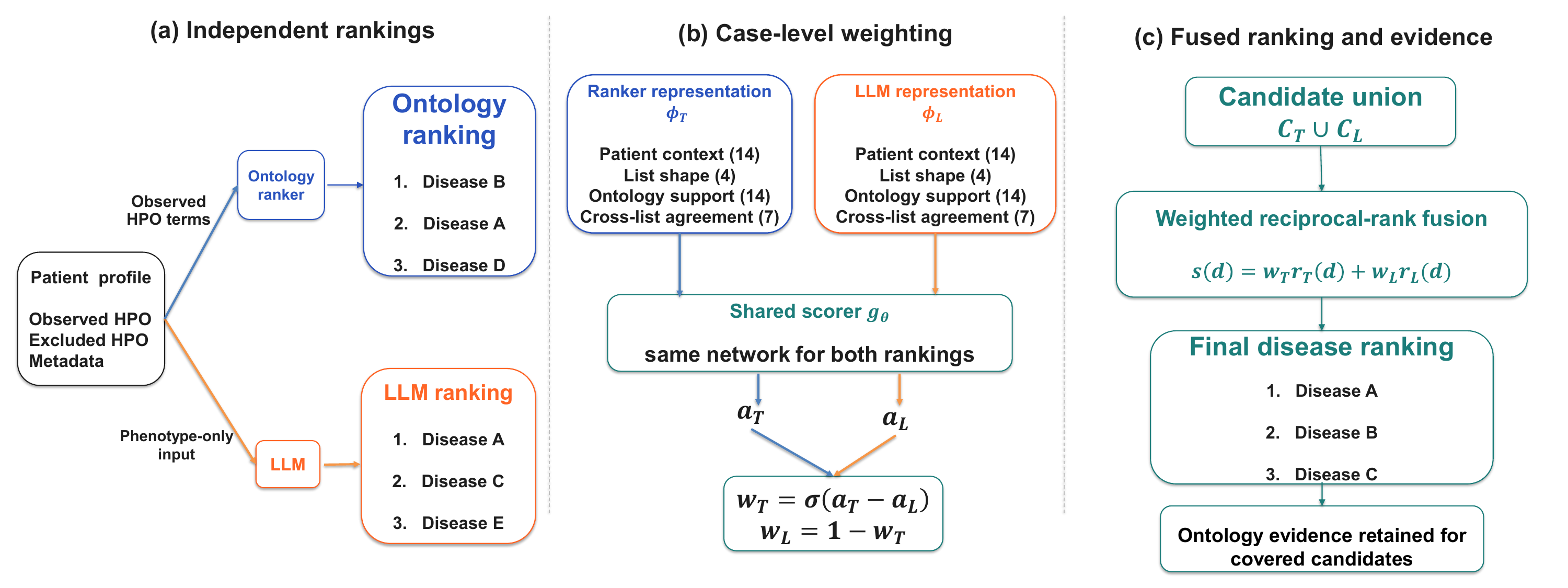}
\caption{Fusion proceeds in three stages. (a) The ontology ranker and the LLM independently
construct disease rankings from the patient profile. (b) A shared scorer maps their case-level
representations to weights. (c) Weighted reciprocal ranks combine the candidate union, while
candidates covered by the ranker retain their ontology evidence.}
\label{fig:fusion}
\end{figure*}

Figure~\ref{fig:fusion} summarizes the model. The ontology ranker and the LLM first construct
separate disease rankings from the available phenotype information. A shared scorer then evaluates
how each ranking behaves on the current case and assigns case-specific weights before their
candidates are combined.

\subsection{Independent Rankings and a Shared Representation}

The ranker compares the patient's observed HPO findings with curated disease profiles. We retain
its first 100 diseases, including their scores, as $C_T$. The LLM reads the
observed and explicitly excluded findings together with the available demographic context, and
produces at most 10 free-text diagnoses, whose names we map to Online Mendelian Inheritance in Man
(OMIM) identifiers with a fixed lexicon to obtain $C_L$. The fused candidate set is $C_T\cup C_L$.
Thus an LLM diagnosis outside the stored ontology
prefix can enter the final ranking; a disease outside both lists cannot.

The resulting lists cannot be combined through their native outputs: the ontology ranker assigns a
score to each candidate, whereas the LLM supplies an ordering without a comparable confidence
scale. We therefore retain only the rank information when the two lists are combined. Let $T$ and
$L$ denote the ontology ranker and the LLM, and let $C_e$ be the candidates returned by system $e$.
For $e\in\{T,L\}$, let $\operatorname{rank}_e(d)$ denote the one-based position of disease $d$ in
$C_e$. We define $r_e(d)$, the rank score assigned to disease $d$ by system $e$, as

\begin{equation}
\label{eq:rankscore}
r_e(d)=
\begin{cases}
1/(\kappa+\operatorname{rank}_e(d)), & d\in C_e,\\
0, & d\notin C_e.
\end{cases}
\end{equation}
The shared nonnegative constant $\kappa$ controls how sharply the score falls down the list:
$\kappa=0$ gives ordinary reciprocal rank, and larger values flatten the decay. We use $\kappa=0$
on Phenopacket Store and $\kappa=60$ on RAMEDIS. This transformation places both lists on the same
scale without treating either system's native output as a calibrated probability.

To estimate how reliable system $e$ is for the current case, we encode its output in a
39-dimensional vector $\phi_e$. The vector contains four groups of signals, each answering a
different question. \emph{Patient context} (14 features) describes how much diagnostic information
the case provides through the number, rarity, and specificity of its observed and excluded HPO
findings, together with the availability of age, onset, and sex. \emph{List shape} (4 features)
captures how decisive the system's differential is through its length and how sharply its leading
candidates stand out from the rest. Because the LLM supplies no native confidence scores, its
candidate at rank $r$ receives the score $1/r$ for these features. \emph{Ontology support} (14 features)
measures how well the leading candidates fit the patient through phenotype similarity,
likelihood-ratio evidence, and the amount of curated phenotype knowledge available for those
diseases. Finally, the \emph{cross-list relationship} (7 features) records whether the two systems
support the same diagnoses, using overlap among their leading candidates and the position that each
system assigns to the other's candidates.

The 14 patient-context features and the four top-$k$ agreement measures ($k=1,3,5,10$) have the
same values in $\phi_T$ and $\phi_L$; the remaining 21 describe the system being scored or how the
other system ranks its candidates. For the leading diseases in either list, we compute the same
HPO-based similarity, likelihood-ratio, and annotation-coverage features. A disease in the LLM's
top 10 is therefore evaluated against the patient's findings and its curated disease profile even
when it falls outside the ontology ranker's retained top 100. On Phenopacket Store, these
calculations use the LOPO-corrected profiles. Model identity, architecture, and parameter count are
never features.
Appendix~\ref{app:features} provides the complete feature definitions.

\subsection{Learning to Combine the Rankings}

We apply the same scoring network $g_\theta$ to the ontology vector $\phi_T$ and the LLM vector
$\phi_L$. The network has two hidden layers, with 48 and 24 units, and produces one scalar for each
component:
\begin{equation}
a_e=g_\theta(\phi_e), \qquad e\in\{T,L\}.
\end{equation}
Only the difference between $a_T$ and $a_L$ affects the fusion. We convert it into two weights:
\begin{align}
w_T &= \sigma(a_T-a_L),
& w_L &= 1-w_T,
\end{align}
where $\sigma$ is the logistic sigmoid. The higher-scoring component receives more influence;
equal scores give both rankings weight $1/2$. The weights are recomputed for every patient.

For every disease $d$ in $C_T\cup C_L$, we compute
\begin{equation}
s(d)=w_T r_T(d)+w_L r_L(d),
\end{equation}
and sort the candidates by $s(d)$. A component contributes zero when the disease is absent from its
list. We train $g_\theta$ with listwise cross-entropy over the candidate union, using the gold
diagnosis to supervise the fused ranking directly; no label specifies which component to trust.

Sharing $g_\theta$ imposes an exchangeable rule. Swapping $(\phi_T,r_T)$ with $(\phi_L,r_L)$
exchanges $w_T$ and $w_L$ and leaves $s(d)$ unchanged for every disease. The final combination uses
ranks, while features derived from native scores are normalized within each list; positive affine
rescaling of a component's scores therefore leaves the fused ranking unchanged.

\subsection{Training}

For the Phenopacket Store transfer experiment, when evaluating a target LLM $L^\star$, we exclude
its rankings and those of every model built on the same backbone. The remaining models provide the
training rankings.

Each training example pairs one case with the ranking produced by one training LLM. It contains the
case's fixed LOPO ontology ranking, that LLM's ranking, and the gold diagnosis. A case therefore
contributes one example for each available training LLM. Because the fusion model can only reorder
diseases in $C_T\cup C_L$, we use an example for training only when this union contains the gold
diagnosis. The same case may appear in several examples with an unchanged patient record and
ontology ranking but a different LLM ranking; no case crosses the publication-disjoint data splits.

Within each training LLM, we sample eligible examples without replacement and allocate the fixed
totals of 7,024 training and 1,094 validation rows as evenly as possible across the available
models. This holds the labelled budget constant across target families. The target family is
removed before sampling, feature standardization, and early stopping;
model identity is never a feature. At test time, each case is paired once with $L^\star$ and passes
through the unchanged map without updating $\theta$ or using target-model labels. A gold disease
outside the candidate union is counted as an error. The experiment therefore tests whether ranking
behavior transfers beyond the model lineage from which it was learned.

\section{Experimental Setup}
\label{sec:setup}

Our primary corpus is Phenopacket Store 0.1.27~\cite{phenopacketstore2025}, containing 10,377
cases from 1,733 source publications and 780 gold diseases. Cases sharing a publication
are kept in the same split, yielding 7,029/1,102/2,246 cases over
1,187/181/365 publications. Ranking experiments require at least one observed
HPO term and a gold disease in the fixed candidate space, leaving
7,024/1,094/2,227 eligible cases in the three splits. No system receives information outside a
closed phenotype-only view comprising canonical labels for observed and explicitly excluded HPO
terms, age, onset, sex, and a hashed case identifier. Each component uses the fields it supports:
Phenomizer receives observed HPO terms, whereas the LLM also receives explicit exclusions and
available demographics. Disease labels, genes, variants, source identifiers, titles, and archive
paths are excluded and checked by a planted-answer negative control. We additionally evaluate the fusion
framework on 624 RAMEDIS cases released with RareBench~\cite{rarebench2024}. RAMEDIS covers 74
inborn errors of metabolism and provides observed HPO findings but not excluded findings, age,
onset, sex, or source-publication identifiers. It therefore uses the ordinary HPOA ranking and a
five-fold evaluation in which identical phenotype profiles remain in the same fold.

The primary ontology system is Phenomizer~\cite{phenomizer2009}, run through the Jackson Laboratory
reference implementation at commit 1dda137 and the published \texttt{phenol}
1.3.3 jars. It scores mean best-match information content and ranks by the empirical
$p$-value produced by its own null model. Its hard-coded query-size cap and tie handling affect
absolute Recall@1, so our provenance estimand is the paired change under the same executable;
Appendix~\ref{app:refimpl} describes the implementation and its sensitivity analyses. We
also evaluate the bare semantic-similarity equation,
three additional editable ontology rankers, and stock LIRICAL 2.4.1 under both ordinary and
exact-LOPO data profiles. Exomiser motivates the comparison but is not a like-for-like component:
its benchmarked clinical mode combines phenotype and variant evidence, while its phenotype-only
mode natively ranks genes rather than the diseases evaluated here~\cite{exomiser2014}.
The
LLM pool contains Qwen2.5-7B-Instruct, Llama3-OpenBioLLM-8B, MedGemma-27B-text-it,
HuatuoGPT-3-8B, Baichuan-M2-32B, MedGemma-4B, Med42-8B, and OpenBioLLM-70B. Each receives the same
bounded prompt and returns at most 10 diagnoses; thinking is disabled where supported.
For each target LLM, the fusion gate is fitted on 7,024 rows after excluding its complete backbone
family. We group OpenBioLLM-8B, OpenBioLLM-70B, and Med42-8B as Llama-3; both MedGemma models as
Gemma-3; and, conservatively, Qwen2.5 and the Qwen3-based HuatuoGPT as one Qwen lineage.
Baichuan is a singleton family. Every arm follows the fixed-budget sampling protocol above, and all
gate results average 5 random seeds. On RAMEDIS, the same 39-feature shared scorer is fitted within
each training fold, selected on a separate validation fold, and evaluated only on the held-out
fold; every reported patient prediction is out of fold. Appendix~\ref{app:ext} gives the exact
split and regularization settings.

Fusion baselines include the two component systems, reciprocal-rank fusion (RRF)~\cite{rrf2009},
Borda-fuse and Bayes-fuse~\cite{aslam2001metasearch}, ProbFuse~\cite{lillis2006probfuse},
and CombMNZ~\cite{foxshaw1994}. We also fit a fixed weight using labels from the target LLM, a
deliberate advantage over transfer without target-model labels. Three learned controls test whether
supervision alone explains the result. Logistic and MLP routers select one component ranking for
each case, while an asymmetric MLP predicts a continuous fusion weight from the two concatenated
component descriptions. The MLP controls have 3,073 parameters, closely matching
the shared scorer's 3,121, and receive exactly the same observable features. All
three are trained with the same family exclusions and labelled budgets as our method. We report Recall@$k$ and MRR, with Recall@1 as the primary
diagnostic endpoint. Learned-system point estimates average five seeds, whose standard deviations
are reported separately. Difference intervals use 2,000 paired percentile-bootstrap resamples,
clustered by source publication on Phenopacket Store and by gold disease on the external corpus.
Each resample averages the same five frozen predictions, so these intervals quantify cohort
sampling rather than optimization variation. Full system versions, prompts, name normalization,
tie policies, and compute are given in the appendix.

\section{Results}
\label{sec:results}

We first measure how publication-source overlap changes ontology ranking. We then evaluate fusion
under the corrected protocol and examine its transfer, architecture, candidate coverage, and
candidate-level ontology evidence.

\subsection{Publication-Source Overlap}
\label{sec:howmuch}

Publication-source overlap is common enough to affect the benchmark materially. Among the
10,348 cases whose gold disease is represented in HPOA, 74.6\% have at least one
gold-disease annotation citing the publication from which the case was constructed. On average,
33.6\% of the gold profile is supported only by that publication; for
19.6\% of cases, this is true of the entire profile.

\begin{table*}[t]\centering\small
\fitwidth{%
\begin{tabular}{lrrr}
\toprule
Ranker & Uncorrected & LOPO & Decrease (pp) [95\% CI] \\
\midrule
Phenomizer & 0.4481 & 0.1217 & 32.64 [26.06, 40.15] \\
Resnik & 0.6484 & 0.2591 & 38.93 [30.59, 47.24] \\
LIRICAL 2.4.1 & 0.4868 & 0.1724 & 31.43 [24.82, 38.72] \\
\bottomrule
\end{tabular}
}
\caption{Recall@1 before and after publication-level LOPO on the 2,227 Phenopacket Store test
cases.}
\label{tab:ranker}
\end{table*}

Across all evaluable cases, Phenomizer Recall@1 falls from 0.4481 to
0.1217 (Table~\ref{tab:ranker}), a paired decrease of 32.64 points
[26.06, 40.15]. The result is not specific to
Phenomizer's statistical layer: applying LOPO to the underlying semantic-similarity score reduces
Recall@1 from 0.6484 to 0.2591, and every tie policy preserves the direction and order of
magnitude. Ties have their largest effect on the bare score, where source-exclusive annotations can
place several candidates at the score ceiling. The empirical $p$-value reduces this ambiguity, but
does not remove the effect of overlap. Full implementation and tie-policy checks are reported in
Appendix~\ref{app:refimpl}.

The same intervention produces nearly the same change in an independently implemented clinical
tool. With the LIRICAL executable and all settings held fixed, substituting the filtered HPOA lowers
Recall@1 from 0.4868 to 0.1724, a paired decrease of
31.43 points [24.82, 38.72]. All 739 cases for which the filter removes
no relation retain byte-identical rankings. The effect is therefore neither an artifact of our
semantic-similarity implementation nor a between-group comparison.

HPOA provenance stops at the publication and cannot assign a relation to one patient within a
multi-case paper. Of the 1,733 source publications, 666 contribute exactly
one phenopacket to the release and 1,067 contribute more than one. Restricting the
test analysis to the 135 cases from the former group, Phenomizer Recall@1
falls from 0.5556 to 0.3481, a
20.74-point decrease [14.07, 28.15]. Only one of these cases loses
its entire gold profile; among the remaining 134, the decrease remains
20.15 points [13.43, 27.61]. This restriction
removes ambiguity between benchmark patients represented from the same paper, while the claim
remains publication-level because a paper may contain patients not represented in the store.

The single-case analysis removes ambiguity about which benchmark patient supplied the annotations,
but a second explanation remains: perhaps any comparable deletion from the correct disease would
cause the same loss. We therefore construct a within-disease control. For each source-exclusive
annotation, we remove instead an independently supported annotation matched for information content,
ontology depth, HPO branch, and contribution to the disease profile. Appendix~\ref{app:matched}
describes the matching procedure.

\begin{table}[t]\centering\small\setlength{\tabcolsep}{3pt}
\fitcol{%
\begin{tabular}{lrrrr}
\toprule
Intervention or cohort & $n$ & before & after & decrease (pp) [95\% CI] \\
\midrule
\multicolumn{5}{l}{\it Matched deletion} \\
Case-source relations & 597 & 0.4238 & 0.1943 & 22.95 [15.10, 32.63] \\
Matched independent relations & 597 & 0.4238 & 0.4305 & -0.67 [-2.39, 0.79] \\
\addlinespace
\multicolumn{5}{l}{\it Gold profile after LOPO} \\
Retains annotations & 1,717 & 0.3722 & 0.1578 & 21.43 [15.54, 28.62] \\
Emptied & 510 & 0.7039 & 0.0000 & 70.39 [60.18, 79.74] \\
\bottomrule
\end{tabular}
}
\caption{Matched-deletion and gold-profile survival analyses. Decreases are paired Recall@1
differences in percentage points; brackets give publication-clustered 95\% confidence intervals.}
\label{tab:provenancechecks}
\end{table}

Table~\ref{tab:provenancechecks} reports this matched control together with a complementary
profile-survival analysis. The upper panel tests generic deletion burden. On the same 597 cases, removing case-source
annotations lowers Recall@1 from 0.4238 to 0.1943, a decrease of 22.95 points
[15.10, 32.63]. Removing equally many matched annotations produces a decrease of
-0.67 points [-2.39, 0.79]. This contrast shows that annotation count,
information content, depth, branch, and profile contribution do not explain the source-deletion
effect. It deliberately does not match agreement with the patient query: that agreement is the
path by which a paper can supply both the test findings and the relations that retrieve them.
Provenance and case-specific semantic alignment are therefore part of the same shortcut mechanism,
not separately identified causal effects.

The lower panel distinguishes weakened ranking signal from the degenerate case in which LOPO
removes the gold profile altogether. Among the 1,717 cases that retain independently supported
gold-disease annotations, LOPO reduces Recall@1 from 0.3722 to 0.1578, a 21.43-point decrease. In
the remaining 510 cases, Recall@1 falls from 0.7039 to zero because the gold disease is no longer
described in HPOA. The all-case estimate in Table~\ref{tab:ranker} measures the ranker's dependence
on the coupled resources; the retained-profile estimate isolates its
nondegenerate ranking effect.

The decrease also grows with the fraction of the gold profile supported only by the case
publication, while uncorrected accuracy is flat across the same strata
(Appendix~\ref{app:dose}). All primary Phenopacket Store fusion experiments use the LOPO ontology
ranking.

\subsection{Diagnostic Accuracy}

\begin{table*}[t]\centering\small\setlength{\tabcolsep}{5pt}
\fitwidth{%
\begin{tabular}{llrrrrr}
\toprule
Held-out family & Target LLM & LLM & RRF & CombMNZ & fixed$^\dagger$ & ours \\
\midrule
Llama-3 & OpenBioLLM-8B & 0.020 & 0.114 & 0.121 & 0.126 & \textbf{0.130}\,\tiny{($\pm$ 0.001)} \\
 & OpenBioLLM-70B & 0.095 & 0.083 & 0.115 & 0.123 & \textbf{0.181}\,\tiny{($\pm$ 0.002)} \\
 & Med42-8B & 0.094 & 0.112 & 0.157 & 0.145 & \textbf{0.183}\,\tiny{($\pm$ 0.003)} \\
\addlinespace
Gemma-3 & MedGemma-4B & 0.092 & 0.135 & 0.178 & 0.139 & \textbf{0.199}\,\tiny{($\pm$ 0.001)} \\
 & MedGemma-27B & 0.106 & 0.090 & 0.121 & 0.132 & \textbf{0.190}\,\tiny{($\pm$ 0.000)} \\
\addlinespace
Qwen-2.5/3 & Qwen2.5-7B-Instruct & 0.139 & 0.115 & 0.181 & 0.145 & \textbf{0.229}\,\tiny{($\pm$ 0.002)} \\
 & HuatuoGPT-3-8B & 0.133 & 0.111 & 0.179 & 0.185 & \textbf{0.227}\,\tiny{($\pm$ 0.001)} \\
\addlinespace
Baichuan & Baichuan-M2-32B & 0.179 & 0.116 & 0.160 & 0.137 & \textbf{0.264}\,\tiny{($\pm$ 0.002)} \\
\midrule
Macro mean & & 0.107 & 0.109 & 0.151 & 0.142 & \textbf{0.200}\,\tiny{($\pm$ 0.001)} \\
\bottomrule
\end{tabular}
}
\caption{Recall@1 on Phenopacket Store with the target backbone family held out. Our results are
means and standard deviations over five seeds; $\dagger$ uses target-LLM labels, whereas our
method does not.}
\label{tab:arms}
\end{table*}

Phenopacket Store provides the cleanest test of transfer to a new model lineage. Across all
8 targets and 4 held-out families, fusion has the highest Recall@1 point estimate against
Phenomizer, the target LLM, RRF, CombMNZ, and a fixed weight fitted with the target model's own
labels (Table~\ref{tab:arms}). Its macro-average is 0.2002 ($\pm$0.0007), compared with
0.1515 for CombMNZ, the strongest of the five published rank-fusion rules. Their paired macro
difference is $+$4.88 points [3.12, 6.77]. The gate therefore transfers through observable ranking
behavior without seeing outputs from the target backbone family.

The improvement extends beyond the first position. Macro Recall@5 and MRR are
0.2937 and 0.2461, compared with 0.2124 and 0.1689 for
Phenomizer and 0.1747 and 0.1393 for the LLMs.

Results are stable to ontology candidate depth. At $K=100$, the gold disease occurs in the
ontology prefix for 46.6\% of test cases and in the test union for
55.1\% of all combinations of test case and target LLM. Complete refits at
$K=10$ and $K=50$ obtain 0.2004 and 0.2008 Recall@1, compared with
0.2002 at $K=100$; replacing the stored prefix at inference with the exact
8,553-disease ordering changes the result by only -0.001 points.

The improvement is not created by the cases whose gold-disease profile is emptied by LOPO. Among
the 1,717 cases with a nonempty independently supported profile, the frozen LOPO
gate raises Recall@1 from 0.1578 to 0.2597, a paired gain of
$+$10.19 points [2.65, 18.04]. We also repeat the complete
family-held-out training protocol using the ordinary, uncorrected Phenomizer ranking and its
corresponding evidence features. Recall@1 rises from 0.4481 to 0.5261,
a gain of $+$7.79 points [1.73, 14.55]. The gate exceeds Phenomizer
for all 8 target models in both checks. LOPO changes how the benchmark
comparison should be interpreted, but the fusion gain does not depend on its lower baseline.

The LLM prompt includes explicitly excluded findings, whereas Phenomizer ranks only the observed
findings. To separate complementarity from this input difference, we replace Phenomizer with the
LOPO naive-Bayes likelihood-ratio ranker, which consumes both observed and explicitly excluded
findings, and repeat the complete family-held-out protocol. Fusion raises its Recall@1 from
0.1361 to 0.2115, a paired gain of 7.54 points
[1.87, 14.12], and exceeds both components in all 8 target
arms. The gain therefore persists when the ontology component has access to the same negative
phenotype information as the LLM.

We further test the practical setting in which a recent frontier model is added after the gate has
been trained. The gate is fitted once on the complete training and validation rankings from the eight
open LLMs, while DeepSeek-V4-Flash is queried only for the held-out test cases. Without any
DeepSeek-labelled example for training, model selection, or feature standardisation,
Recall@1 increases from 0.1217 for Phenomizer and 0.1657 for DeepSeek
alone to 0.2176 for fusion, averaged over five seeds. These are gains of
9.59 and 5.19 percentage points over the two
components, respectively. Thus the gate can improve an unseen model outside its training pool,
rather than only interpolate among the LLMs on which it was learned.

\begin{table*}[t]\centering\small\setlength{\tabcolsep}{4pt}
\fitwidth{%
\begin{tabular}{lrrrrrr}
\toprule
Target LLM & Phenomizer & LLM & RRF & Borda-fuse & CombMNZ & ours \\
\midrule
Baichuan-M2-32B & 0.1554 & 0.2837 & 0.3942 & 0.3846 & 0.3253 & \textbf{0.4006}\,\tiny{($\pm$ 0.0032)} \\
HuatuoGPT-3-8B (thinking disabled) & 0.1554 & 0.2548 & 0.3221 & 0.3237 & 0.3253 & \textbf{0.3272}\,\tiny{($\pm$ 0.0033)} \\
Llama3-OpenBioLLM-8B & 0.1554 & 0.0224 & 0.1571 & 0.1587 & \textbf{0.1619} & 0.1577\,\tiny{($\pm$ 0.0008)} \\
Llama3-OpenBioLLM-70B & 0.1554 & 0.3061 & 0.4103 & 0.4054 & 0.3333 & \textbf{0.4128}\,\tiny{($\pm$ 0.0008)} \\
Med42-8B & 0.1554 & 0.3301 & 0.3734 & 0.3654 & 0.3606 & \textbf{0.3833}\,\tiny{($\pm$ 0.0006)} \\
MedGemma-4B & 0.1554 & 0.2019 & \textbf{0.3702} & 0.3654 & 0.3654 & 0.3692\,\tiny{($\pm$ 0.0008)} \\
MedGemma-27B-text-it & 0.1554 & \textbf{0.4599} & 0.4295 & 0.3894 & \textbf{0.4599} & 0.4183\,\tiny{($\pm$ 0.0014)} \\
Qwen2.5-7B-Instruct & 0.1554 & 0.1971 & \textbf{0.3926} & 0.3862 & 0.2740 & 0.3888\,\tiny{($\pm$ 0.0031)} \\
\midrule
Macro mean & 0.1554 & 0.2570 & 0.3562 & 0.3474 & 0.3257 & \textbf{0.3573} \\
\bottomrule
\end{tabular}
}
\caption{Recall@1 on RAMEDIS. RRF, Borda-fuse, and CombMNZ require no training. Ours is
evaluated by five-fold profile-grouped cross-validation. Learned results are means and standard
deviations over five seeds.}
\label{tab:externalarms}
\end{table*}

RAMEDIS provides an independently assembled, information-limited test of the same fusion principle.
The gate is refitted within the RAMEDIS folds, so this experiment tests fusion on a second corpus
rather than parameter transfer from Phenopacket Store. Across the
eight LLMs, our method raises Phenomizer's macro Recall@1 from 0.1554 to
0.3573, a gain of 20.18 points
[4.90, 33.79], and exceeds the mean standalone LLM by
10.02 points [2.79, 15.46]
(Table~\ref{tab:externalarms}). It improves on Phenomizer in every arm and on both component
systems in 7 of 8. Its macro average is the highest
in the table, and it is the strongest listed method for
4 individual LLMs.

RAMEDIS also delineates when learning provides additional value beyond rank aggregation. The
corpus contains 624 cases from 74 metabolic diseases and supplies observed HPO findings, but not
the excluded findings or demographic context used by several patient-context features. Under this
restricted input contract, RRF reaches 0.3562 and our method reaches 0.3573, a difference of
0.11 points [-0.32, 0.46]. The two are therefore statistically indistinguishable on RAMEDIS,
showing that parameter-free aggregation can suffice when the available case description is sparse.

\subsection{Gate Architecture and Features}

We compare the shared scorer with supervised controllers under identical inputs, family
exclusions, labelled budgets, and test cases.
The logistic and MLP routers learn when to select the ontology ranking or the LLM ranking. The
asymmetric MLP instead predicts a continuous weight from their concatenated descriptions, using
the same listwise objective as our gate. Unlike the shared scorer, these controls may attach a
different meaning to the same feature according to which component produced it.

\begin{table}[t]\centering\small\setlength{\tabcolsep}{3pt}
\fitcol{%
\begin{tabular}{lrrr}
\toprule
Control & dim. & Recall@1 & decrease (pp) \\
\midrule
Full shared MLP & 39 & 0.2002\,\tiny{($\pm$ 0.0007)} & n/a \\
$-$ ontology support & 25 & 0.1780\,\tiny{($\pm$ 0.0032)} & 2.23 [0.89, 3.93] \\
Support + agreement only & 21 & 0.1991\,\tiny{($\pm$ 0.0010)} & 0.12 [-0.10, 0.36] \\
\addlinespace
Logistic router & 78 & 0.1956\,\tiny{($\pm$ 0.0008)} & 0.46 [0.05, 0.90] \\
MLP router & 78 & 0.1959\,\tiny{($\pm$ 0.0016)} & 0.43 [-0.02, 1.02] \\
Asymmetric fusion MLP & 78 & 0.1914\,\tiny{($\pm$ 0.0023)} & 0.88 [0.10, 2.23] \\
\bottomrule
\end{tabular}
}
\caption{Feature ablations and learned controls under family holdout. Values are macro Recall@1,
reported as mean and standard deviation over five seeds; decreases are relative to the full shared
MLP, with publication-clustered 95\% confidence intervals.}
\label{tab:featureablation}
\end{table}

In the five-seed main experiment, the shared scorer reaches 0.2002 Recall@1, compared
with 0.1956, 0.1959, and 0.1914 for logistic routing, MLP routing,
and asymmetric fusion (Table~\ref{tab:featureablation}). These paired cohort intervals average the
five seed predictions and do not measure optimization variation. We therefore repeat the shared
scorer and both routers under 30 matched initializations. Its mean advantage is
0.43 points [0.38, 0.49] over logistic routing and
0.35 points [0.27, 0.42] over MLP routing, with positive paired
differences in 30/30 and
29/30 seeds, respectively
across the matched initializations. The symmetric architecture is consequently a
small, reproducible refinement, not the primary source of the fusion gain.

Within the shared architecture, the decisive signal is evidence about the proposed diseases.
Removing ontology support produces by far the largest decrease, from 0.2002 to
0.1780 Recall@1. A 21-feature scorer retaining only
ontology support and cross-list agreement reaches 0.1991; its Phenopacket Store
difference from the full contract is not detectable. Thus ontology support and agreement account
for nearly all of the measurable Phenopacket Store gain. We retain the complete contract as one
representation for corpora that provide different amounts of case context.

A linear shared scorer reaches 0.1881, 3.66 points above CombMNZ
but 1.21 points below the MLP. Most of the difference from CombMNZ therefore
comes from supervised behavioral and ontology evidence; nonlinear interactions and the symmetric
scorer add smaller refinements.

Appendix~\ref{app:results} reports the remaining provenance analyses, ranking metrics, feature
ablations, candidate-depth checks, and rank-fusion comparisons.

\subsection{Candidate Coverage and Ontology Evidence}

The LOPO-corrected Phenopacket Store benchmark reveals substantial complementarity between the two diagnostic
systems. An oracle that accepts a correct top prediction from either component exceeds the stronger
component by 10.33 Recall@1 points, and the LLM ranks the gold disease first in
0.183 of Phenomizer's errors. After LOPO, the gold disease has a median of
34 direct HPOA annotations, compared with 12 across the
8,553-disease candidate vocabulary. When a gold profile is sparse, the ontology ranker has less
curated evidence on which to act, leaving more room for an independently generated LLM
differential to add the diagnosis.

For 90.8\% of correct fused predictions, the disease occurs in Phenomizer's stored
top 100, where its matched HPO terms and $p$-value are available for review. This is candidate-level
ontology evidence, not an interpretation of the gate's complete decision path. Fusion is not
restricted to that prefix: 18.0\% of all fused top predictions, including
9.2\% of the correct ones, originate outside it. Outside-prefix promotions are correct in only
10.3\%, compared with 52.7\% for promotions from
Phenomizer ranks two through ten. Candidate expansion can therefore recover diagnoses absent from
the stored prefix, but those predictions are less reliable and carry no stored ontology evidence.

Transfer also depends on the LLM returning an informative, normalizable ranking. MedGemma-27B
includes the gold disease in 33.4\% of its lists, but places it first in only
31.8\% of those cases. Qwen2.5-7B has lower list recall
(30.0\%) but commits to the gold disease more often once it is present
(46.3\%). OpenBioLLM-8B produces no normalized list for
955 test cases.

To distinguish failures of name normalization from unusable model output, two clinical experts
independently reviewed a blinded sample of 400 failed names or empty parses,
100 from each of 4 models. Among the
190 randomly sampled unmatched occurrences, the experts classified
between 22.1\% and 23.2\% as mapper misses,
whereas between 48.9\% and 52.6\% were invalid or hallucinated. Category agreement was
86.0\% (Cohen's $\kappa=0.782$). Inserting every
candidate-space OMIM mapping supplied by either expert repaired
between 41 and 53 names and changed between 5 and 9 predicted top labels,
but neither changed a single Top-1 correctness outcome (0.00 Recall@1
points). Thus mapper misses are a real interface loss, but they do not explain the audited Top-1
results. Sampling and category definitions are given in Appendix~\ref{app:setup}.

\section{Discussion and Conclusion}

The provenance analysis changes how retrospective evaluations of knowledge-based diagnosis should
be read. When a test case and an inference-time knowledge base are curated from the same paper,
measured accuracy reflects both ranking quality and the reuse of document-specific evidence. LOPO
exposes this dependence, while the matched-deletion and retained-profile analyses show that it is
not explained by generic annotation loss or by cases whose gold profile disappears. For evaluations
built from curated cases and curated external resources, shared source provenance is therefore part
of benchmark design rather than incidental metadata.

The corrected evaluation also changes the role of the LLM. An independently generated differential
can broaden the candidate set, while the ontology ranker supplies structured phenotype evidence for
the candidates it covers. Family-held-out training and the test-only DeepSeek arm show that the gate
can exploit this complementarity without labels from the target model family. The gain under the
ordinary, uncorrected protocol shows that fusion is not an artifact of the lower LOPO baseline, and
RAMEDIS extends the result to an independently assembled disease cohort and marks a boundary of the
learned gate. With observed phenotypes but no exclusions or demographic context, the gate and RRF
are statistically indistinguishable. In this sparse-input setting, combining the rankings matters,
but learning their weights provides no detectable additional gain.

Together, these results support a hybrid role for LLMs in phenotype-based diagnosis. The LLM need
not replace the ontology tool: behavior-based fusion improves its ranking while preserving
candidate-level ontology evidence for most correct predictions. Because the gate depends on
observable ranking behavior rather than model identity, a newly introduced LLM can enter the
pipeline without target-model labels. Auditing resource provenance and combining complementary
diagnostic evidence offer a cleaner and more adaptable path than treating structured tools and LLMs
as competing alternatives.

\section*{Limitations}

The provenance analysis concerns Phenopacket Store and HPOA, where both cases and annotations record
their source publications. LOPO removes this recorded overlap from the ontology side; it does not
audit LLM pretraining data or imply that the same overlap is widespread in other rare-disease
benchmarks. Our experiments also address phenotype-only disease ranking rather than Exomiser's
variant-aware clinical mode.

The fusion model requires disease names that can be mapped to a shared ontology. Family-held-out and
test-only experiments support transfer across the models studied, but not every future model or
clinical setting. These experiments evaluate diagnostic ranking and candidate-level evidence on
public research corpora; they do not establish clinical validity or support autonomous diagnosis.

\section*{Ethics Statement}

We use de-identified case descriptions from public research corpora and collect no new patient
data. Two clinical experts reviewed de-identified model outputs for the disease-name audit; they
did not diagnose or recommend treatment for any real patient. Our experiments evaluate
retrospective disease ranking rather than clinical validity, and none of the evaluated systems
should be used for autonomous diagnosis. AI assistants supported implementation and language
editing; the authors verified the study design, analysis, claims, and reported results.

\bibliographystyle{acl_natbib}
\bibliography{custom}

\appendix

\section{Reproducibility Details}
\label{app:setup}

This appendix follows the evaluation pipeline in the order in which a case is processed. We first
describe the corpus and the phenotype-only input, then document the LLM and ontology rankers, the
fusion experiments, and the external evaluation.

\subsection{Corpus, Inputs, and Evaluation}

\paragraph{Corpus and cohorts.}
Phenopacket Store 0.1.27 contains 10,377 cases from 1,733 publications. Keeping cases from the same
publication together gives 7,029 training, 1,102 validation, and 2,246 test cases. Ranking requires
at least one observed HPO term and a gold disease in the 8,553-disease candidate space, leaving
7,024, 1,094, and 2,227 eligible cases, respectively. The provenance analysis additionally requires
the gold disease to be represented in HPOA, which holds for 10,348 cases. Table~\ref{tab:corpusstats}
summarizes the corpus.

\begin{table}[t]\centering\small
\fitcol{%
\begin{tabular}{lr}
\toprule
Cases & 10,377 \\
Training / validation / test & 7,029 / 1,102 / 2,246 \\
Ranking-eligible split & 7,024 / 1,094 / 2,227 \\
Distinct diseases (OMIM) & 780 \\
Distinct causal genes & 699 \\
Distinct source publications & 1,733 \\
Observed HPO terms & 92,068 \\
Excluded HPO terms & 132,795 \\
Median HPO terms per case & 15 \\
Cases with onset or age & 88.0\% \\
Candidate diseases (OMIM) & 8,553 \\
\bottomrule
\end{tabular}
}
\caption{Phenopacket Store 0.1.27 corpus and evaluation cohorts.}
\label{tab:corpusstats}
\end{table}

\paragraph{Phenotype-only input.}
Phenopackets include fields that can reveal the answer, including disease and interpretation
records, external references, record identifiers, and archive paths. We therefore construct each
input from a fixed set of permitted fields: HPO CURIEs and their canonical labels from the pinned
\texttt{hp.json}, sex, normalized age or onset, and a hashed case identifier. No other text enters
the model input.

The verifier scans the resulting input for gold disease identifiers, gene symbols, HGNC
identifiers, variant expressions, publication identifiers, and archive paths. As a negative
control, it inserts a gold disease name into a phenotype label and confirms that the audit detects
it. It also verifies that no publication crosses the data splits. Some lexical overlap is intrinsic
to the ontology: in 34.7\% of cases, an observed phenotype label and the gold disease name share a
content word. We retain these valid HPO labels and present them identically to every system.

\paragraph{Systems and evaluation.}
The editable ontology systems are the reference Phenomizer implementation
(Appendix~\ref{app:refimpl}), the Resnik score, symmetric best-match-average Resnik, and two
LIRICAL-style naive-Bayes likelihood-ratio rankers that differ in whether they use excluded
phenotypes. We also run the unmodified LIRICAL 2.4.1 executable against each filtered HPOA snapshot.
We report Recall@$k$ and MRR. Differences on Phenopacket Store use 2,000 paired bootstrap samples
clustered by source publication. RAMEDIS does not provide case-source publications, so its paired
bootstrap samples are clustered by gold disease. Intervals for learned systems are computed from
predictions averaged across five frozen seeds; seed variation is reported separately.

\subsection{LLM Inference and Disease-Name Mapping}

\paragraph{Models.}
The open-model pool comprises Qwen2.5-7B-Instruct, Llama3-OpenBioLLM-8B,
MedGemma-27B-text-it, HuatuoGPT-3-8B, Baichuan-M2-32B, MedGemma-4B, Med42-8B, and
OpenBioLLM-70B. Each model receives the same phenotype-only input and is asked to return up to 10
ranked disease names. Thinking is disabled where supported. DeepSeek-V4-Flash is evaluated only at test time
through SiliconFlow: its rankings are passed to the frozen gate, but it contributes no training or
validation examples.

\paragraph{Prompt and decoding.}
Every LLM receives the following system message:
\begin{quote}\small
You are a clinical geneticist performing phenotype-only differential diagnosis of rare Mendelian
disease. You are given a patient's Human Phenotype Ontology findings. Reply with exactly 10
candidate diagnoses, most likely first, one per line, formatted as ``\textless rank\textgreater.
\textless disease name\textgreater''. Use standard OMIM/Mondo disease names. No explanations, no
other text.
\end{quote}
Each user message gives sex and age or onset when available, followed by all observed HPO terms and
any explicitly excluded terms. It ends with the instruction ``List the 10 most likely rare genetic
diagnoses, ranked.'' Apart from the structured demographic fields, the variable clinical text
consists only of canonical HPO labels. Local models use their native chat templates and greedy
decoding with temperature 0, one completion, a 4,096-token context limit, and at most 384 generated
tokens. DeepSeek uses the same messages and decoding settings through the API. The verifier flags
outputs containing explicit reasoning markers. The parser accepts numbered lines matching
\texttt{\textless rank\textgreater[.)] \textless disease name\textgreater} and retains the first
10 names.

\paragraph{Disease-name mapping.}
Surface forms are mapped to OMIM identifiers with a fixed lexicon built from the candidate diseases
and the names and exact synonyms of Mondo~\cite{mondo2022} classes with an OMIM cross-reference.
The mapper rejects matches that change an explicitly numbered disease subtype.

We audit 400 unmatched items, with 100 drawn from Baichuan-M2-32B, MedGemma-4B,
Llama3-OpenBioLLM-8B, and Qwen2.5-7B-Instruct. For each model, the sample includes up to 20 outputs
from which the parser recovered no disease name. The remaining quota is divided equally between
randomly sampled unmatched occurrences and unique unmatched surface forms.
Two clinical experts independently classify each item as an invalid or hallucinated name, a mapper
miss, a granularity mismatch, a valid non-OMIM disease, or a formatting failure. Model identity,
case identifier, and gold diagnosis are hidden. We retain both judgments. For the random-occurrence
stratum, the main text therefore reports the range between the two experts' category estimates and
evaluates each expert's proposed OMIM mappings separately.

\subsection{Ontology Rankers and Provenance Controls}
\label{app:refimpl}

\paragraph{Reference Phenomizer implementation.}
We run the public implementation at
\url{https://github.com/TheJacksonLaboratory/Phenomiser} at commit
1dda137, the last commit that retains the $p$-value module. This version uses
\url{https://github.com/monarch-initiative/phenol} 1.3.3 (tag
\texttt{v1.3.3}, commit a7f2f66). We use the published Maven artifacts rather than a local build.
Before execution, the verifier checks their SHA-1 digests against the served values:
b42430be\allowbreak 0f60815e\allowbreak f3f2e0d8\allowbreak b5b9d81a\allowbreak f4c3f42c,
096ac16a\allowbreak 46861c63\allowbreak 2d357d46\allowbreak 02fdd78c\allowbreak 12dc785e, and
a18da0be\allowbreak 3ae6eef1\allowbreak 5ab91098\allowbreak fb24c16e\allowbreak 11ab17d8.

The reference software performs its own null sampling, ranking, and comparison over 8,553 candidate
diseases, using 100,000 Monte Carlo samples. We supply the query term sets and information-content
values. Three of the 10,377 cases contain no canonical HPO term after input filtering and cannot be
ranked because \texttt{phenol} has no size-zero null distribution. The same cases are absent under
both evaluation protocols. Because Monte Carlo sampling is stochastic, Table~\ref{tab:mc} reports
four complete runs.

\paragraph{Query-size cap.}
The reference code constructs null distributions only for queries of at most 10 terms. A longer
query is scored with all of its terms but is compared with the size-10 null distribution. We retain
this native behavior for the main results. Table~\ref{tab:cap} reports a sensitivity analysis using
our diagnostic variant with the cap removed.

\paragraph{Native ranking and ties.}
Phenomizer converts semantic similarity into an empirical $p$-value against random queries for
each candidate. It orders candidates by ascending $p$-value and then by descending similarity
$S$. When both values are equal, stable downstream sorts preserve the implementation's original
iteration order. Across 4,454 test-case and protocol combinations, 626 rank-1 predictions
(14.1\%) fall inside an unresolved block of this kind. The mean block size is 1.41 and the maximum
is 22. We therefore report the native list order together with tie-averaged and worst-case
sensitivities; only the native ordering is used in the fusion experiments.

\begin{table}[t]\centering\small\setlength{\tabcolsep}{3pt}
\fitcol{%
\begin{tabular}{lrrr}
\toprule
run & Recall@1 uncorrected & Recall@1 LOPO & accuracy ratio \\
\midrule
1 (the run reported throughout) & 0.4481 & 0.1217 & 3.683 \\
2 & 0.4499 & 0.1307 & 3.443 \\
3 & 0.4522 & 0.1226 & 3.689 \\
4 & 0.4504 & 0.1266 & 3.557 \\
\bottomrule
\end{tabular}
}
\caption{Variation across four Phenomizer runs with 100,000 Monte Carlo samples.}
\label{tab:mc}
\end{table}

\begin{table}[t]\centering\small\setlength{\tabcolsep}{3pt}
\fitcol{%
\begin{tabular}{lrr}
\toprule
& uncorrected & LOPO \\
\midrule
Recall@1, capped (reference) & 0.4481 & 0.1217 \\
Recall@1, uncapped diagnostic variant & 0.4688 & 0.1441 \\
\midrule
\multicolumn{3}{l}{\emph{cases over the cap (n=681)}} \\
\quad capped (reference) & 0.6931 & 0.1307 \\
\quad uncapped (ours) & 0.7753 & 0.2012 \\
\multicolumn{3}{l}{\emph{cases at or under it (n=1,546)}} \\
\quad capped (reference) & 0.3402 & 0.1177 \\
\quad uncapped (ours) & 0.3338 & 0.1190 \\
\bottomrule
\end{tabular}
}
\caption{Recall@1 with the native 10-term cap and with the diagnostic uncapped variant.}
\label{tab:cap}
\end{table}

\paragraph{Effect of the statistical layer.}
\label{app:null}

The empirical null correction explains why the reference implementation differs from the bare
semantic score in Table~\ref{tab:ranker}. The median gold disease has 52 annotations before LOPO
and 34 afterward, compared with 12 for a candidate drawn from the full space. A correction for
profile richness may therefore attenuate genuine signal together with annotation-density bias.

The Monte Carlo estimate also has limited resolution at the top of the ranking. Sorting the saved
reference outputs by $p$-value alone yields 0.0965 Recall@1 before LOPO and 0.0427
afterward; 77.7\% of uncorrected cases place rank one inside a block tied on $p$-value alone. The
similarity criterion resolves most of these primary-key ties, leaving the 14.1\% unresolved rate
reported above.

\paragraph{Exact LOPO for LIRICAL.}
For each test publication, we create a temporary LIRICAL data directory containing every pinned
2.4.1 resource except \texttt{phenotype.hpoa}. In its replacement, a relation between an OMIM
disease and an HPO term is removed only when the test publication is its sole cited source. All
cases from the same publication use this filtered resource with LIRICAL's stock phenotype-only
command. The runner verifies the JAR, HPOA, split, and baseline hashes, stores resumable per-case
results, and removes the temporary directory. The run covers the 2,227 ranking-eligible cases from
364 publications.
For 739 cases, no relation is removed and the filtered and unfiltered rankings are identical.

\paragraph{Matched independent deletion.}
\label{app:matched}

The matched control begins with the 1,440 test cases for which exact LOPO removes at least one
direct annotation from the gold disease. If LOPO removes $k$ annotations, the control pool contains
direct annotations of the same disease with recorded support outside the case publication. A case
is retained only when this pool contains at least $k$ annotations. This leaves 597 cases containing
793 removed annotations; the other 843 cases do not have enough eligible controls.

We obtain a minimum-cost bipartite assignment without replacement. The matching variables are
full-HPO information content, ontology depth, the logarithm of the annotation's unique contribution
to the inherited disease profile, and its major HPO branch. Patient findings and ranking outcomes
are not used. Of the 793 matched pairs, 73.9\% differ by at most one information-content unit and
89.3\% by at most one ontology level. The standardized mean differences are \(-0.009\) for
information content and \(+0.018\) for depth.

Each retained query is ranked under three HPOA conditions using the same base null samples and Java
implementation: LOPO across all diseases, source-specific deletion from the gold disease only, and
matched independent deletion from the gold disease. The unedited baseline is identical across the
three conditions. The first two obtain Recall@1 of 0.1943 on this cohort, whereas matched deletion
obtains 0.4305. The confidence interval in Table~\ref{tab:provenancechecks} resamples source
publications, keeping cases from the same paper together.

\subsection{Fusion Models and Baselines}
\label{app:features}

Table~\ref{tab:features} lists the 39 features used to represent each component ranking, in their
implementation order. Patient-level features are copied into both component vectors. List
geometry, ontology support, and directional cross-list features are computed separately for the
ontology ranker and the LLM. Rarity is information content, defined as $-\log$ of the fraction of
diseases annotated with a finding; specificity is depth below the HPO root.

\begin{table*}[p]\centering\footnotesize
\fitwidth{%
\begin{tabular}{p{0.30\linewidth}p{0.62\linewidth}}
\toprule
\multicolumn{2}{l}{\textbf{Patient-level features} (14, shared by both vectors)} \\
\midrule
observed finding count & number of HPO findings recorded as present \\
excluded finding count & number of HPO findings recorded as absent \\
log observed count & $\log(1+{}$observed finding count$)$ \\
excluded fraction & excluded $/$ (observed $+$ excluded) \\
mean observed rarity & mean information content of observed findings \\
maximum observed rarity & largest observed information content \\
minimum observed rarity & smallest observed information content \\
observed rarity spread & standard deviation of observed information content \\
mean observed specificity & mean ontology depth of observed findings \\
maximum observed specificity & largest ontology depth among observed findings \\
mean excluded rarity & mean information content of excluded findings \\
onset available & 1 if age of onset is recorded, otherwise 0 \\
age available & 1 if age is recorded, otherwise 0 \\
sex available & 1 if sex is recorded, otherwise 0 \\
\addlinespace
\multicolumn{2}{l}{\textbf{Component-ranking statistics} (4)} \\
\midrule
log list length & $\log(1+{}$number of returned candidates$)$ \\
top-score margin & first score minus second after rescaling scores to $[0,1]$ \\
top-10 score entropy & entropy of the rescaled top-10 scores \\
score spread & standard deviation of the rescaled list scores \\
\addlinespace
\multicolumn{2}{l}{\textbf{Ontology support for leading candidates} (14)} \\
\midrule
rank-1 match score & semantic similarity of the case to the first candidate \\
mean match score, top 3 & mean semantic similarity over the first three candidates \\
mean match score, top 10 & mean semantic similarity over the first ten candidates \\
maximum match score, top 10 & largest semantic similarity among the first ten \\
match-score spread, top 10 & standard deviation among the first ten \\
rank of maximum match score & position of the largest top-10 semantic similarity \\
rank-1 evidence score & likelihood ratio for the first candidate \\
mean evidence score, top 3 & mean likelihood ratio over the first three candidates \\
mean evidence score, top 10 & mean likelihood ratio over the first ten candidates \\
maximum evidence score, top 10 & largest likelihood ratio among the first ten \\
evidence-score spread, top 10 & standard deviation among the first ten \\
rank of maximum evidence score & position of the largest top-10 likelihood ratio \\
rank-1 profile size & $\log(1+{}$HPO annotations for the first candidate$)$ \\
mean profile size, top 10 & mean log annotation count over the first ten candidates \\
\addlinespace
\multicolumn{2}{l}{\textbf{Cross-ranking agreement} (7)} \\
\midrule
reciprocal other-list rank, top 1 & reciprocal rank of this list's first candidate in the other list; 0 if absent \\
mean reciprocal other-list rank, top 3 & mean reciprocal rank for this list's first three candidates \\
mean reciprocal other-list rank, top 10 & mean reciprocal rank for this list's first ten candidates \\
top-1 agreement & 1 if both lists have the same first candidate, otherwise 0 \\
top-3 Jaccard agreement & intersection over union of the two top-3 candidate sets \\
top-5 Jaccard agreement & intersection over union of the two top-5 candidate sets \\
top-10 Jaccard agreement & intersection over union of the two top-10 candidate sets \\
\addlinespace
\bottomrule
\end{tabular}
}
\caption{The 39 features used to represent each component ranking.}
\label{tab:features}
\end{table*}

\paragraph{Gate optimization.}
The shared scorer uses ReLU hidden layers of 48 and 24 units. We standardize each feature with the
mean and standard deviation computed over both component vectors in the source training rows,
adding $10^{-6}$ to each standard deviation. AdamW uses a learning rate of $2\times10^{-3}$,
weight decay of $10^{-4}$, and batches of 512. Training runs for at most 100 epochs and stops after
12 validation epochs without improvement in listwise cross-entropy. The fused rank scores are
multiplied by 8 inside this training loss.

\paragraph{Learned controls.}
The two routing controls are trained only on source-family cases for which the ontology and LLM
top-1 predictions differ in correctness. The routing label specifies which complete ranking to
select; cases for which both predictions have the same outcome provide no routing preference. The
logistic router is linear, while the MLP router has nearly the same parameter count as the shared
scorer. The asymmetric fusion control uses the same MLP capacity but predicts a continuous mixture
weight under the listwise objective. Both MLP controls concatenate the two 39-dimensional feature
vectors, giving 78 inputs. Feature standardization, early stopping, and checkpoint selection use
source families only. The selected models are applied unchanged to the held-out family.

\paragraph{Rank-fusion baselines.}
\label{app:pubfusion}
We compare with RRF~\cite{rrf2009}, Borda-fuse and Bayes-fuse~\cite{aslam2001metasearch},
ProbFuse~\cite{lillis2006probfuse}, and CombMNZ~\cite{foxshaw1994}. Every method receives the same
rankings, candidate pools, cases, and metrics. RRF, Borda-fuse, ProbFuse, and CombMNZ follow
\texttt{ranx} 0.3.21~\cite{bassani2022ranx}; Bayes-fuse follows the cited likelihood-ratio rule and
assigns neutral evidence to an absent candidate. Trained baselines use the same source rows as the
gate and no labels from the held-out model family.

\subsection{Additional Evaluation}

\paragraph{Source-overlap sensitivity.}
\label{app:dose}

The dose analysis asks whether the LOPO effect grows with the fraction of the gold disease profile
supported only by the case publication. It includes the 7,717 ranking-eligible cases with source
overlap and uses 2,000 bootstrap samples clustered by source publication.

\paragraph{Cross-corpus overlap.}
\label{app:external}
We apply the same PMID join to RareArena~\cite{rarearena2025}, excluding publications shared with
Phenopacket Store before estimating overlap. RareArena uses Orphanet disease identifiers, so cases
are mapped to unique OMIM equivalents before they are joined to HPOA. We also check whether each
source PMID occurs anywhere in the 8,958-publication HPOA curation pool, a bound that does not
depend on disease mapping. As a mapping control, we send the Phenopacket Store disease identifiers
through the same OMIM to Orphanet to OMIM procedure and confirm that the strong overlap signal
remains detectable.

\paragraph{RAMEDIS.}
\label{app:ext}

RAMEDIS is the 624-patient inborn-error-of-metabolism split from
RareBench~\cite{rarebench2024}. We use five-fold StratifiedGroupKFold cross-validation, stratifying
by gold disease and grouping identical observed-HPO profiles so that duplicate profiles cannot
cross folds. In each outer split, one fold is held out for testing, one is used for checkpoint
selection, and the remaining three train the gate. Training and validation examples are built
separately from their respective patient folds and use rankings from all eight LLMs, including the
target LLM; evaluation pairs each held-out test patient only with the corresponding target LLM
ranking. We repeat the procedure for five seeds. RAMEDIS uses
$\kappa=60$ in Equation~\ref{eq:rankscore}. The validation fold also selects
$\alpha\in\{0,0.125,0.25,0.5,0.75,1\}$ and shrinks the learned ontology weight to
$1/2+\alpha(w_T-1/2)$. Neither choice uses test labels, predictions, or feature statistics.

\subsection{Compute and Resources}
\label{app:compute}

All ontology rankers, provenance interventions, and fusion models run on CPU. A reference
Phenomizer run with 100,000 Monte Carlo samples takes 1,752.5 seconds on 92 threads. Local LLM
inference uses vLLM on an NVIDIA RTX 6000 Ada with 48 GB memory; MedGemma-27B uses FP8. Total local
GPU time is under six hours. The shared scorer has 3,121 parameters.

The resources are Phenopacket Store 0.1.27 under BSD-3-Clause, HPO and HPOA under the HPO license,
Mondo under CC-BY 4.0, LIRICAL 2.4.1, Phenomiser commit 1dda137, \texttt{phenol} 1.3.3,
\texttt{ranx} 0.3.21, RareBench under its stated terms, and the licenses distributed with the open
models.

\section{Complementary Results}
\label{app:results}

We first examine how publication-source overlap varies across cases and corpora. We then
report complementary ranking metrics, ablations, sensitivity to candidate depth, and comparisons
with published fusion rules.

\subsection{Publication-Source Overlap}

\begin{table}[H]\centering\small
\fitcol{%
\begin{tabular}{lr}
\toprule
Statistic & Value \\
\midrule
Cases with publication-source overlap & 74.6\% \\
Gold-disease annotations that are source-exclusive (mean) & 33.6\% \\
\quad median & 12.3\% \\
Cases with an \emph{entirely} source-exclusive gold profile & 19.6\% \\
Query terms covered by source-exclusive gold annotations (mean) & 37.7\% \\
\bottomrule
\end{tabular}
}
\caption{Publication-source overlap between Phenopacket Store and HPOA ($n=10{,}348$).}
\label{tab:mech}
\end{table}

\begin{table}[H]\centering\small\setlength{\tabcolsep}{3pt}
\fitcol{%
\begin{tabular}{lrrl}
\toprule
Corpus & analysed cases & source overlap & relation to Phenopacket Store \\
\midrule
Phenopacket Store & 10,348 & 74.6\% & primary \\
RareArena RDS benchmark & 2,645 & 0.76\% & independently assembled \\
RareArena RDC benchmark & 1,481 & 0.81\% & independently assembled \\
LIRICAL benchmark & 385 & 51.95\% & 85.71\% same paper/disease \\
\bottomrule
\end{tabular}
}
\caption{Publication-source overlap with HPOA in three rare-disease resources. For LIRICAL, the
last column gives the fraction of its cases with the same source PMID and gold disease as a
Phenopacket Store case.}
\label{tab:crosscorpus}
\end{table}

\begin{figure}[H]\centering
\includegraphics{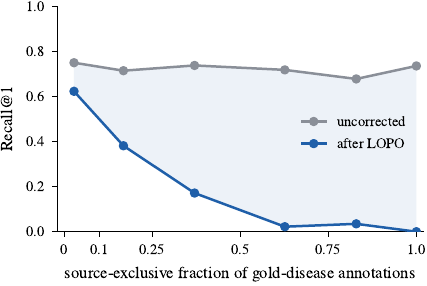}
\caption{Recall@1 by the source-exclusive fraction of the gold disease profile. The six points are
placed at the mean fraction within each interval.}
\label{fig:dose}
\end{figure}

\subsection{Fusion Robustness}

\begin{table}[h!]\centering\small\setlength{\tabcolsep}{4pt}
\fitcol{%
\begin{tabular}{lrrrr}
\toprule
System & Recall@1 & Recall@5 & Recall@10 & MRR \\
\midrule
Phenomizer & 0.1217 & 0.2124 & 0.2434 & 0.1689 \\
LLM macro & 0.1072 & 0.1747 & 0.2075 & 0.1393 \\
RRF & 0.1094 & 0.2589 & 0.3333 & 0.1789 \\
Target-labelled fixed & 0.1415 & 0.2877 & 0.3453 & 0.2141 \\
Family-held-out gate & \textbf{0.2002} & \textbf{0.2937} & \textbf{0.3502} & \textbf{0.2461} \\
\bottomrule
\end{tabular}
}
\caption{Macro ranking performance on Phenopacket Store. Gate values average five seeds; the LLM
row averages eight target models.}
\label{tab:rankingmetrics}
\end{table}

\begin{table}[t]\centering\small\setlength{\tabcolsep}{3pt}
\fitcol{%
\begin{tabular}{lrrr}
\toprule
Control & dim. & Recall@1 & decrease (pp) \\
\midrule
\multicolumn{4}{l}{\it Feature and scorer controls} \\
Full shared MLP & 39 & 0.2002\,\tiny{($\pm$ 0.0007)} & n/a \\
$-$ patient context & 25 & 0.1984\,\tiny{($\pm$ 0.0010)} & 0.18 [-0.01, 0.42] \\
$-$ list shape & 35 & 0.2001\,\tiny{($\pm$ 0.0012)} & 0.02 [-0.13, 0.17] \\
$-$ ontology support & 25 & 0.1780\,\tiny{($\pm$ 0.0032)} & 2.23 [0.89, 3.93] \\
$-$ cross-list agreement & 32 & 0.1974\,\tiny{($\pm$ 0.0007)} & 0.28 [0.00, 0.63] \\
Support + agreement only & 21 & 0.1991\,\tiny{($\pm$ 0.0010)} & 0.12 [-0.10, 0.36] \\
Linear shared scorer & 39 & 0.1881\,\tiny{($\pm$ 0.0020)} & 1.21 [0.41, 2.10] \\
\addlinespace
\multicolumn{4}{l}{\it Alternative learned controllers} \\
Logistic router & 78 & 0.1956\,\tiny{($\pm$ 0.0008)} & 0.46 [0.05, 0.90] \\
MLP router & 78 & 0.1959\,\tiny{($\pm$ 0.0016)} & 0.43 [-0.02, 1.02] \\
Asymmetric fusion MLP & 78 & 0.1914\,\tiny{($\pm$ 0.0023)} & 0.88 [0.10, 2.23] \\
\bottomrule
\end{tabular}
}
\caption{Feature ablations and learned controls under family holdout. Parentheses give seed
standard deviations; decreases are relative to the full model, with 95\% confidence intervals.}
\label{tab:featureablationfull}
\end{table}

\begin{table}[t]\centering\small\setlength{\tabcolsep}{3pt}
\fitcol{%
\begin{tabular}{lrrr}
\toprule
System & Recall@1 & shared $-$ control (pp) & wins \\
\midrule
Shared scorer & 0.1997 $\pm$ 0.0010 & n/a & n/a \\
Logistic router & 0.1954 $\pm$ 0.0010 & 0.43 [0.38, 0.49] & 30/30 \\
MLP router & 0.1962 $\pm$ 0.0015 & 0.35 [0.27, 0.42] & 29/30 \\
\bottomrule
\end{tabular}
}
\caption{Paired results over 30 initializations. Brackets give seed-level 95\% confidence intervals.}
\label{tab:optimizationvariance}
\end{table}

\begin{table}[t]\centering\small\setlength{\tabcolsep}{4pt}
\fitcol{%
\begin{tabular}{lrrr}
\toprule
Candidate cutoff $K$ & ontology coverage & union coverage & Recall@1 \\
\midrule
10 (refitted) & 24.3\% & 43.7\% & 0.2004 \\
50 (refitted) & 36.6\% & 50.5\% & 0.2008 \\
100 (refitted) & 46.6\% & 55.1\% & 0.2002 \\
8,553 (inference only) & 100.0\% & 100.0\% & 0.2002 \\
\bottomrule
\end{tabular}
}
\caption{Gold-disease coverage and Recall@1 at different ontology candidate cutoffs. The final row
expands inference for the gate fitted at $K=100$.}
\label{tab:candidatedepth}
\end{table}

\begin{table}[t]\centering\small
\fitcol{%
\begin{tabular}{lrrr}
\toprule
Rule & list order & tie-averaged & worst case \\
\midrule
RRF & 0.1094 & 0.1009 & 0.0885 \\
Borda-fuse & 0.1022 & 0.1010 & 0.0968 \\
Bayes-fuse & 0.1199 & 0.0669 & 0.0331 \\
ProbFuse & 0.1201 & 0.0788 & 0.0538 \\
CombMNZ$^\ddagger$ & 0.1515 & 0.1338 & 0.1051 \\
\midrule
Family-held-out gate (ours) & \textbf{0.2002} & n/a & n/a \\
\bottomrule
\end{tabular}
}
\caption{Phenopacket Store Recall@1 for published rank-fusion rules, averaged over eight target
LLMs. The three columns apply the native list order, average tied ranks, or assign the least
favorable rank within each tie. $^\ddagger$ marks the \texttt{ranx} score convention.}
\label{tab:pub}
\end{table}

\end{document}